\documentclass{bmvc2k}

\title{Efficient Bird Eye View Proposals for 3D Siamese Tracking}

\addauthor{Jesus Zarzar}{jesusalejandro.zarzartorano@kaust.edu.sa}{1}
\addauthor{Silvio Giancola}{silvio.giancola@kaust.edu.sa}{1}
\addauthor{Bernard Ghanem}{bernard.ghanem@kaust.edu.sa}{1}

\addinstitution{
 Image and Video Understanding Laboratory\\
 King Abdullah University of Science and Technology\\
 Thuwal, KSA
}
\runninghead{Zarzar, Giancola, Ghanem}{BEV Proposals for 3D Tracking}

\usepackage{booktabs}

\def\eg{\emph{e.g}\bmvaOneDot}
\def\Eg{\emph{E.g}\bmvaOneDot}
\def\etal{\emph{et al}\bmvaOneDot}

\begin{document}

\maketitle

\newcommand{\mysection}[1]{\vspace{3pt}\noindent\textbf{#1.}}
\newcommand{\TODO}[1]{\textcolor{red}{\textbf{\textit{[TODO] #1}}}}
\newcommand{\BG}[1]{\textcolor{red}{\textbf{\textit{[BG] #1}}}}
\newcommand{\JZ}[1]{\textcolor{red}{\textbf{\textit{[JZ] #1}}}}
\newcommand{\SG}[1]{\textcolor{red}{\textbf{\textit{[SG] #1}}}}
\newcommand{\sota}{state-of-the-art\xspace}
\newcommand{\Sota}{State-of-the-art\xspace}
\newcommand{\Table}[1]{Table~\ref{tab:#1}}
\newcommand{\Figure}[1]{Figure~\ref{fig:#1}}
\newcommand{\Equation}[1]{Equation~\eqref{eq:#1}}
\newcommand{\Section}[1]{Section~\ref{sec:#1}}
\newcommand{\job}[1]{\textcolor{red}{\textbf{\textit{[job-id] #1}}}}

\newcommand{\AVOD}{Ku~\etal~\cite{ku2018joint}\xspace}
\newcommand{\MVTD}{Chen~\etal~\cite{chen2017multi}\xspace}
\newcommand{\SiamRPN}{Li~\etal~\cite{Li_2018_CVPR}\xspace}
\newcommand{\BEV}{Bird Eye View\xspace}

\makeatletter
\DeclareRobustCommand\onedot{\futurelet\@let@token\@onedot}
\def\@onedot{\ifx\@let@token.\else.\null\fi\xspace}

\def\eg{\emph{e.g}\onedot} \def\Eg{\emph{E.g}\onedot}
\def\ie{\emph{i.e}\onedot} \def\Ie{\emph{I.e}\onedot}
\def\cf{\emph{c.f}\onedot} \def\Cf{\emph{C.f}\onedot}
\def\etc{\emph{etc}\onedot} \def\vs{\emph{vs}\onedot}
\def\wrt{w.r.t\onedot} \def\dof{d.o.f\onedot}
\def\etal{\emph{et al}\onedot}
\makeatother

%%%%%%%%% ABSTRACT
\begin{abstract}
Tracking vehicles in LIDAR point clouds is a challenging task due to the sparsity of the data and the dense search space. The lack of structure in point clouds impedes the use of convolution filters usually employed in 2D object tracking. In addition, structuring point clouds is cumbersome and implies losing fine-grained information. As a result, generating proposals in 3D space is expensive and inefficient. In this paper, we leverage the dense and structured Bird Eye View (BEV) representation of LIDAR point clouds to efficiently search for objects of interest. We use an efficient Region Proposal Network and generate a small number of object proposals in 3D. Successively, we refine our selection of 3D object candidates by exploiting the similarity capability of a 3D Siamese network. We regularize the latter 3D Siamese network for shape completion to enhance its discrimination capability. Our method attempts to solve both for an efficient search space in the BEV space and a meaningful selection using 3D LIDAR point cloud. We show that the Region Proposal in the BEV outperforms Bayesian methods such as Kalman and Particle Filters in providing proposal by a significant margin and that such candidates are suitable for the 3D Siamese network. By training our method end-to-end, we outperform the previous baseline in vehicle tracking by $12\% / 18\%$ in Success and Precision when using only 16 candidates.
\end{abstract}

\section{Introduction}

Autonomous vehicles have to understand their environment in order to fulfill their driving task.
Understanding the road can be achieved by different means: an autonomous agent can either follow the human process, \ie sense the world though RGB images, or rely on more advanced sensors, such as LIDAR devices.
On one hand, Convolutional Neural Networks (CNNs) contributed largely to the improvement of 2D object detection, tracking, and segmentation tasks, making the use of RGB images ubiquitous~\cite{ren2015faster,he2016deep,badrinarayanan2017segnet}.
However, hard conditions such low-light environment and low-texture objects could confuse any algorithm as much as a human driver.
On the other hand, LIDAR sensors estimate depth using an active Time-of-Flight principle, which is less sensitive to texture, and more robust low-light condition.
Therefore, LIDARs provide reliable geometrical information which could provide a better understanding of the road, the actors involved and their dynamics than camera relying only on visual information.

\begin{figure*}
    \centering
    \includegraphics[width=\textwidth]{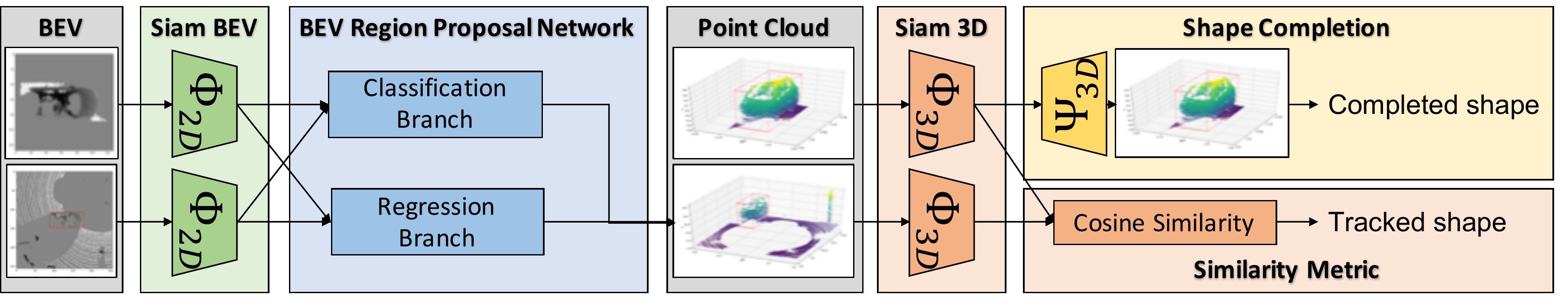}
    \caption{
    Vehicle tracking is performed with a double Siamese network for both 3D point cloud (orange) and 2D Bird Eye View (green) representations of the LIDAR data.
    We leverage an efficient search space using a Region Proposal Network (blue) on top of the BEV Siamese network to generate fast proposals.
    We leverage a shape completion regularization (yellow) on top the 3D Siamese network for a semantic similarity metric.
    As a result, we track vehicles using exclusively LIDAR data and reconstruct a more complete vehicle shape.
    }
    \label{fig:Inference}
    \label{fig:Pooling}
\end{figure*}

Here we focus on 3D vehicle tracking, \ie following the position of a vehicle in a real-time fashion. Such position is defined using a 3D bounding box and initialized in the first frame of a sequence.
Giancola~\etal~\cite{CVPR19} introduced the first 3D Siamese Tracker. 
They solve for a similarity metric in a PointNet encoding and show improvement in 3D Tracking by regularizing the latent space for Shape completion. % to resemble a shape completion latent space.
However, they sample candidates using an \emph{exhaustive search} that includes the ground truth bounding box, which is idealistic yet not realistic.
When Kalman or Particle Filters are used for the search space, their performance drops drastically, indicating room for improvement.

In contrast, we generate efficient bounding box proposals for vehicle tracking in LIDAR point clouds.
An overview of our pipeline is shown in \Figure{Pooling}.
We propose a double Siamese network that encodes both point cloud and a 2D Bird Eye View (BEV) representation.
Bounding box proposals are generated using a Region Proposal Network (RPN) in the 2D BEV representation, successively used to train the 3D Siamese network.
We show that the BEV representation contains enough information to provide good-enough candidates which are successively discarded using the 3D Siamese features for the tracking task.

\mysection{Contributions:}
Our contributions are threefold:
%
% \begin{enumerate}
\textbf{(i)} We provide an novel efficient search space using a Region Proposal Network (RPN) in the Bird Eye View (BEV) space for the task of 3D Vehicle Tracking in Autonomous Driving settings.
\textbf{(ii)} We train a double Siamese network end-to-end so to learn proposals and similarity scores simultaneously.
\textbf{(iii)} We show that using both 2D and 3D Siamese networks and regularizing for shape completion improves performances by $12\% / 18\%$ in both Success and Precision with respect to the state-of-the-art proposed in \cite{CVPR19} while processing fewer candidates.
% \end{enumerate}
\emph{Code will be released upon publication.}

% \newpage
\section{Related Work}

Our work takes insights from Siamese Trackers, 3D Pose Estimation and Search Spaces.

\mysection{Siamese Trackers}
The task of tracking is currently solved using Siamese Trackers to follow the trajectory of objects~\cite{Muller_2018_ECCV}, people~\cite{luber2011people}, vehicles~\cite{Geiger2012KITTI} or visual attributes~\cite{vot_tpami}, given an initial position.
The problem is commonly tackled through \emph{tracking-by-detection}, where an \emph{object representation} is built after the first frame and a \emph{search space} is constructed weighting computational costs versus denser space sampling.
Algorithms generally use visual or shape information provided through images~\cite{vot_tpami}, depth maps~\cite{luber2011people,song2013tracking}, and/or LIDAR point clouds~\cite{Geiger2012KITTI,NuScenes}.
Bertinetto~\etal~\cite{bertinetto2016fully} were the first to introduce a Siamese network to learn a more abstract representation of the object to track, relying on correlation to find its position in the next frame.
Li~\etal~\cite{Li_2018_CVPR} introduced a Region Proposal Network to avoid convolving the whole search space and Zhu~\etal~\cite{Zhu_2018_ECCV} Hard Negative Mining to focus on the most challenging samples during training.
More recently, Li~\etal~\cite{li2018siamrpn++} extended the Siamese network by using ResNet-50 as the backbone \cite{he2016deep}. %, which achieved \sota results in the challenging large-scale TrackingNet dataset~\cite{Muller_2018_ECCV}.
Giancola~\etal~\cite{CVPR19} introduced the first 3D Siamese tracker based solely on point clouds. 
They regularized the 3D latent space for shape completion and showed improved results in the similarity metric they exploit for tracking.
However, they assume an exhaustive search space that guarantees the ground truth sample is included as one of the candidates and disregard the problem of generating candidates.
Due to the continuous 3D space and the inevitable drift that occurs while tracking, a smart search space is necessary.
In our work, we propose an efficient search space leveraging a Region Proposal Network on the BEV space to provide a limited set of good object candidates.

\mysection{Rigid Registration}
In robotics, the 3D pose of an object pose can be estimated using Rigid Registration methods such as ICP. 
Yet, registration methods are prone to local minima and long convergence. While a step of ICP can be performed in a fraction of ms, it may require numerous iterations before convergence.
In contrast, a two-stage proposal pose estimation is suitable for parallel computing.
Giancola~\etal~\cite{CVPR19} can elaborate up to 147 candidates \emph{simultaneously} in a less than $2.0$ ms.
In this work, we out-perform them by using only $16$ candidates, so more objects/candidates can be processed using a single GPU.

\mysection{3D Tracking for Autonomous Driving}
In road settings, tracking is typically reduced to $3$ degrees of freedom using the hypothesis that vehicles lean on the 3D plane created by the road and are rigid objects. 
Vehicle tracking can be solved leveraging 3D Detection \cite{chen2017multi,ku2018joint,Luo_2018_CVPR} in a Multi Object fashion, but will lack of temporal consistency. 
In our work, we target Single Object Tracking to focus exclusively on the task of tracking.

\mysection{Particle/Kalman Filterings}
Kalman and Particle Filters are widely used in robotics~\cite{chen2012kalman} and visual object tracking~\cite{ristic2003beyond,zhang2015structural}.
They fit a distribution based on observations and update that distribution according to the particles' observations.
They embrace a wide literature including
importance sampling \cite{blake1997condensation}, 
non-collapsing particles \cite{musso2001improving},
hierarchical \cite{yang2005fast},
multi-task \cite{zhang2017multi}
learnable \cite{karkus2018particle} particle filters.
In this work, we compare against the particle filtering employed by Giancola~\etal~\cite{CVPR19}.

\mysection{3D Object Proposal}
3D Object Proposals can be interpreted as an alternative to particle filter-based sampling.
Shin \etal~\cite{shin2016object} provide a proposal method based on point cloud clustering. Even though the method provides interesting results, they do not compare on elaboration times which are too slow for tracking purposes.
% Several work focused on object proposal.
Ren~\etal~\cite{ren2015faster} proposed a Region Proposal Network (RPN) for the Fast RCNN object detector. Several works focused on adapting this work for use on 3D data.
%Deep Sliding Shapes
Song~\etal~\cite{song2016deep} proposed Deep Sliding Shapes, an amodal 3D object detection framework for RGB-D frames. 
They find the main orientation of a scene based on a Manhattan Frame module, coupled with an RPN acting on a TSDF representation of the scene to produce 3D proposal BBs.
%MV3D
Chen~\etal~\cite{chen2017multi} proposed MV3D, an object detection network based on RGB images as well as the top and front views of LIDAR point clouds. 
They have a 3D proposal network based on RPN coupled with a region-based fusion network.
%VOXELNET
Zhou~\etal~\cite{zhou2017voxelnet} proposed VoxelNet, an end-to-end learning method for point cloud based 3D object detection. 
They learn features on a voxelization of the space and use a fully-convolutional neural network to output region proposals in 3D.
While these methods provide interesting insights for proposals on different sensors, none of them actually leverage exclusively LIDAR information.
We motivate the efficient LIDAR BEV representation for proposals and the meaningful 3D point cloud representation for similarity estimation in tracking settings.

% \newpage
\section{Methodology}

We propose a double Siamese network that leverages both 2D BEV and 3D point cloud representations for the LIDAR data. 
Searching in BEV images is easier than in point clouds, although prone to error due to the coarse appearance information contained in a 2D BEV representation.
However, the BEV representation is suitable for proposal generation as soon as it provides a good recall.
The fine-grained details contained in the 3D point cloud representation are afterwards used to select the correct candidate among the proposals and refine the search.
An overview of our method is shown in \Figure{Pooling}.

\subsection{LIDAR Frames and Bounding Boxes Representations}

\mysection{3D Tracking Settings}
The input data representation is a set of temporally contiguous LIDAR frames. 
We assume the pose of the object to track to be given in the first frame.
The pose of an object is defined with the smallest 3D bounding box around it, constrained to lie on the horizontal plane of the road.

\mysection{3D Bounding Boxes}
The pose of an object is generally defined using $9$ parameters for its \emph{position}, \emph{orientation} and \emph{dimension}, reduced to $3$ parameters in our 3D Tracking settings.
We restrict the \emph{position} and \emph{orientation} for 3D Tracking to lie on the horizontal plane of the road. 
Also, we hypothesise for rigid object tracking, with constant bounding boxes \emph{dimension} in time, initialized in the first frame.
Note that the rigid object hypothesis holds for 3D Tracking on the KITTI dataset~\cite{Geiger2012KITTI} for the Car, Cyclist and Pedestrian object classes.
Such a 3D bounding box projects onto the 2D BEV space as an oriented rectangle, defined by $3$ parameters: $x$, $z$ and $\alpha$.
There exists a bijection between the 2D and 3D bounding boxes since the scale and the height component of the shape are assumed to be constant along a trajectory.

\mysection{LIDAR Frame Representations}
The raw representation for a LIDAR frame is a sparse 3D \emph{Point Cloud (PC)}.
We build a \emph{BEV} representation for a given frame by projecting its \emph{Point Cloud} as an image of size $H \times W$ with $(N+2)$ channels consisting of $N=1$ vertical slices, a maximum-height and a density map.
The vertical extent of point cloud data is taken to be $[-1, 1]$ meters from the center of the tracked object.
A \emph{model PC} is maintained for the object 3D shape, initialized from the 3D bounding box around the object in the first frame. 
The \emph{model PC} is iteratively updated concatenating the points in the the bounding box prediction after each frame.
A \emph{model BEV} representation is produced using the equivalent BEV representation of the \emph{model PC}, by cropping an area of $[-2.5, 2.5] \times [-2.5, 2.5]$ meters around the \emph{model PC}.
A \emph{search BEV} representation is generated at each frame by cropping an area of $[-5, 5] \times [-5, 5]$ meters around the previously estimated position in the BEV representation.
We use a dimension of $255 \times 255$ and $127 \times 127$ pixels for the the \emph{search BEV} and \emph{model BEV}, respectively, corresponding to a resolution of approximately $0.04$ meters/pixel.

\subsection{Our Novel BEV-3D Siamese Architecture}

Our architecture is composed of
a \emph{BEV Siamese Network} between the model BEV and the search BEV,
a \emph{Region Proposal Network} that produces a set of $C$ candidates bounding boxes in the current frame and 
a \emph{3D Siamese Network} that discriminates between the candidates using a more reliable model PC.

\mysection{BEV Siamese Network}
We propose a BEV Siamese network that encodes the \emph{BEV model} and the \emph{BEV search} space representation.
The backbone $\Phi_{2D}$ is based on AlexNet~\cite{krizhevsky2012imagenet} and pre-trained on ImageNet.
We set the number of input channels for the first convolutional layer to be the number of layers that compose our BEV representation, \ie $N+2$.
We extract a latent feature map of size $6\times6$ for the model BEV as well as a search space of size $22\time22$.
Both features maps have a dimension of $256$.

\mysection{BEV Region Proposal Network}
We adapt the Region Proposal Network (RPN) from SiamRPN~\cite{SiamRPN} to generate proposals for the model BEV.
The RPN is composed of a classification branch and a regression branch which convolve separately the model and search space feature maps from the Siamese network and correlate the search space features with the model features.
The classification branch selects a set of $C$ candidates from the remaining search space of size $17\times17$. 
We leverage $K=5$ different anchors to cover multiple angles in the range of $[-5, 5]$ degrees.
Note that the classification scores are weighted along a cosine window, as a common practice for 2D object tracking.
Inspired by \cite{ren2015faster,Luo_2018_CVPR}, we successively regress the delta position $\delta_x = \frac{x - x_a}{w}$ and $\delta_z  = \frac{z - z_a}{l}$ as the difference between the anchor priors ($x_a$, $z_a$) and the object coordinate ($x$, $z$), in a space normalized by the dimension of the tracked object ($w$, $l$).
The delta position ($\delta_x$, $\delta_z$) are then added to the anchor priors ($x_a$, $z_a$) to generate the refined proposal.

\mysection{3D Siamese Network}
The proposals are converted in 3D candidate bounding boxes to fit with the 3D Point Cloud representation of the LIDAR frame.
The points lying within the 3D candidate bounding boxes are leveraged as \emph{candidate shapes} for the object of interest.
We enforce the \emph{candidate shapes} to contains $2048$ points by either discarding or duplicating points in a uniform random way.
We use the 3D Siamese network introduced by \cite{CVPR19} to classify whether a candidate shape correspond to the \emph{model PC} using their 3D point cloud representations.
This 3D Siamese network $\Phi_{3D}$ consists of $3$ PointNet layers~\cite{qi2017pointnet} which provide a good trade off between representation capabilities and fast processing.

\subsection{Training for the BEV-3D Siamese Network}

The BEV Siamese Network with BEV Region Proposal is trained alongside the 3D Siamese Network, to generate reliable candidate shapes from their BEV representations and further classify their 3D Point Cloud representations.

\mysection{Region Proposal Loss}
We train our Region Proposal Network to classify and regress a set of $17\times17\times5$ anchors along the search space.
Among the $1445$ anchors, we select $48$ anchors for each frame.
From the $48$ anchors, at most $16$ are positive, where positive anchors are defined as having an IoU in the range $]0.5,1]$ with the ground truth.
$16$ more anchors are taken from the pool of anchors having an IoU in the range $]0, 0.5]$ with respect to the ground truth.
The remaining $16$ anchors are taken from the anchors which do not overlap with the ground truth bounding box.
We adopt a Binary Cross Entropy loss for the classification and a Smooth L1 loss for the regression, as shown in \Equation{LossRPN}. 
The classification loss run over the $N_t=48$ selected anchors for a given frame, with $p_\mathbf{x}$ being the classification score for a given anchor $\mathbf{x}$, and $p^*_\mathbf{x}$ its ground truth label.
In contrast, the regression loss run over the $N_p=16$ positive anchors only, with $\delta^\mathbf{x}_{x/z}$ being the regressed delta position for the anchor $\mathbf{x}$, and $\delta^{\mathbf{x}^*}_{x/z}$ its ground truth value.
Both losses are averaged across the total number of examples for the batch.

\begin{equation}
    \begin{split}
        \mathcal{L}_{cls} &= \frac{1}{N_t} \sum_\mathbf{x}{p^*_\mathbf{x} log(p_\mathbf{x}) + (1 - p^*_\mathbf{x})log(1 - p_\mathbf{x})}  \\
        \mathcal{L}_{reg} &= \frac{1}{2 N_p} \sum_\mathbf{x}{\gamma^\mathbf{x}_x + \gamma^\mathbf{x}_z} ,  ~~~ \text{with} ~~\gamma^{\mathbf{x}}_{x/z} = 
        % \gamma^{i}_x = % \left.
          \begin{cases}
          0.5 (\delta^\mathbf{x}_{x/z} - \delta^{\mathbf{x}^*}_{x/z})^2 ,  &\text{if} \quad | \delta^\mathbf{x}_{x/z} - \delta^{\mathbf{x}^*}_{x/z} | < 1 \\
            | \delta^\mathbf{x}_{x/z} - \delta^{\mathbf{x}^*}_{x/z} | - 0.5,  &\text{otherwise}
          \end{cases}
        %   \right
    \end{split}
    \label{eq:LossRPN}
\end{equation}

\mysection{3D Tracking Regression Loss}
We use the $48$ anchors selected by the Region Proposal Network for the 3D tracking loss.
Candidate shapes are defined by the points contained in the anchor boxes in 3D.
The model shape is defined in training as the concatenation of all ground truth shapes along a tracklet.
We learn to regress a cosine similarity between an encoded candidate shape $\Phi_{3D}(\mathbf{x})$ and the encoded model shape $\Phi_{3D}(\mathbf{\hat{x}})$ to their Gaussian distance $\rho \big( d \left( \mathbf{x} , \mathbf{\hat{x}} \right) \big)$ in the 3D search space.
This tracking loss is shown in \Equation{Loss3DTracking}.

\begin{equation}
    \mathcal{L}_{tr} =  \frac{1}{N_t} \sum_\mathbf{x} \Big( CosSim \big(\phi_{3D}(\mathbf{x}), \phi_{3D}(\mathbf{\hat{x}}) \big) -  \rho \big( d \left( \mathbf{x} , \mathbf{\hat{x}} \right) \big) \Big) ^2 
    \label{eq:Loss3DTracking}
\end{equation}

\mysection{3D Completion Loss}
Similar to Giancola~\etal~\cite{CVPR19}, we leverage a shape completion regularization in order to embed shape features within the Siamese latent space.
The Chamfer Loss is used for this purpose between the model shape $\hat{x}$ and its more complete reconstruction, $\tilde{x}$, as per \Equation{Loss3DCompletion}.
The model shape is obtained as the concatenation of all ground truth shapes along a tracklet.

\begin{equation}
    \mathcal{L}_{comp} = 
    \sum_{\mathbf{\hat{x}}_i \in \mathbf{\hat{x}}} \min_{\mathbf{\tilde{x}}_j \in \mathbf{\tilde{x}}} \|\mathbf{\hat{x}}_i - \mathbf{\tilde{x}}_j\|_2^2 + 
    \sum_{\mathbf{\tilde{x}}_j \in \mathbf{\tilde{x}}} \min_{\mathbf{\hat{x}}_i \in \mathbf{\hat{x}}} \|\mathbf{\hat{x}}_i - \mathbf{\tilde{x}}_j\|_2^2
    \label{eq:Loss3DCompletion}
\end{equation}

\mysection{Training details}
We train the region proposal network alongside the 3D Siamese network by minimizing the loss shown in \Equation{LossFinal} using an image-centric sampling approach as is common with RPNs.
In particular, we use 
$\lambda_{cls} = 1e^{-2}$, 
$\lambda_{reg} = 1$, 
$\lambda_{tr} = 1e^{-2}$ and
$\lambda_{comp} = 1e^{-6}$.
During training, we minimize the loss $\mathcal{L}$ using SGD with a momentum of $0.9$. 
We start with a learning rate of $10^{-4}$ which we reduce by a factor of $10$ after each plateau of the validation loss, with a patience of $2$.
Note that the training usually converges after a few epochs since our networks are pretrained.

\begin{equation}
    \mathcal{L} = \lambda_{cls} \mathcal{L}_{cls} + \lambda_{reg} \mathcal{L}_{reg} + \lambda_{tr} \mathcal{L}_{tr} + \lambda_{comp} \mathcal{L}_{comp} 
    \label{eq:LossFinal}
\end{equation}

\mysection{Testing details}
During inference, the pipeline is initialized with the ground truth bounding box for the tracked object in the first frame.
Tracklets are processed in an online fashion looking only at the current and previous frames for each time step.
The model PC is initialized to be the shape enclosed by the ground truth bounding box in the initial frame.
For each tracking step, a search area is cropped around the previous location of the tracked object.
BEV images are created for both the model PC and search PC to be processed by the RPN.
The RPN generates $C$ candidates which are then classified by the 3D Siamese network by comparing them with the model PC.
Finally, the candidate with the best score from the 3D Siamese network is selected as the pose of the tracked object for the current frame.

% \newpage
\section{Experiments}
\label{sec:Experiments}

% In this section, we compare our RPN module for generating proposals over the Kalman and Particle Filters proposed by Giancola~\etal~\cite{CVPR19}.
% Successively, we show the results of the proposal selection method using the 3D Siamese network.
We first provide the results of our search space for cars, and consider cyclist and pedestrian in the latter.
We report the One Pass Evaluation (OPE)~\cite{vot_tpami} metrics from Single Object Tracking defined with the Success as the AUC of the IOU and the Precision as the AUC of the distance between both centers.

\subsection{Main Results}

\begin{figure}[htb]
    \centering
    \begin{minipage}{0.36\linewidth}
    \centering
    \includegraphics[width=\textwidth]{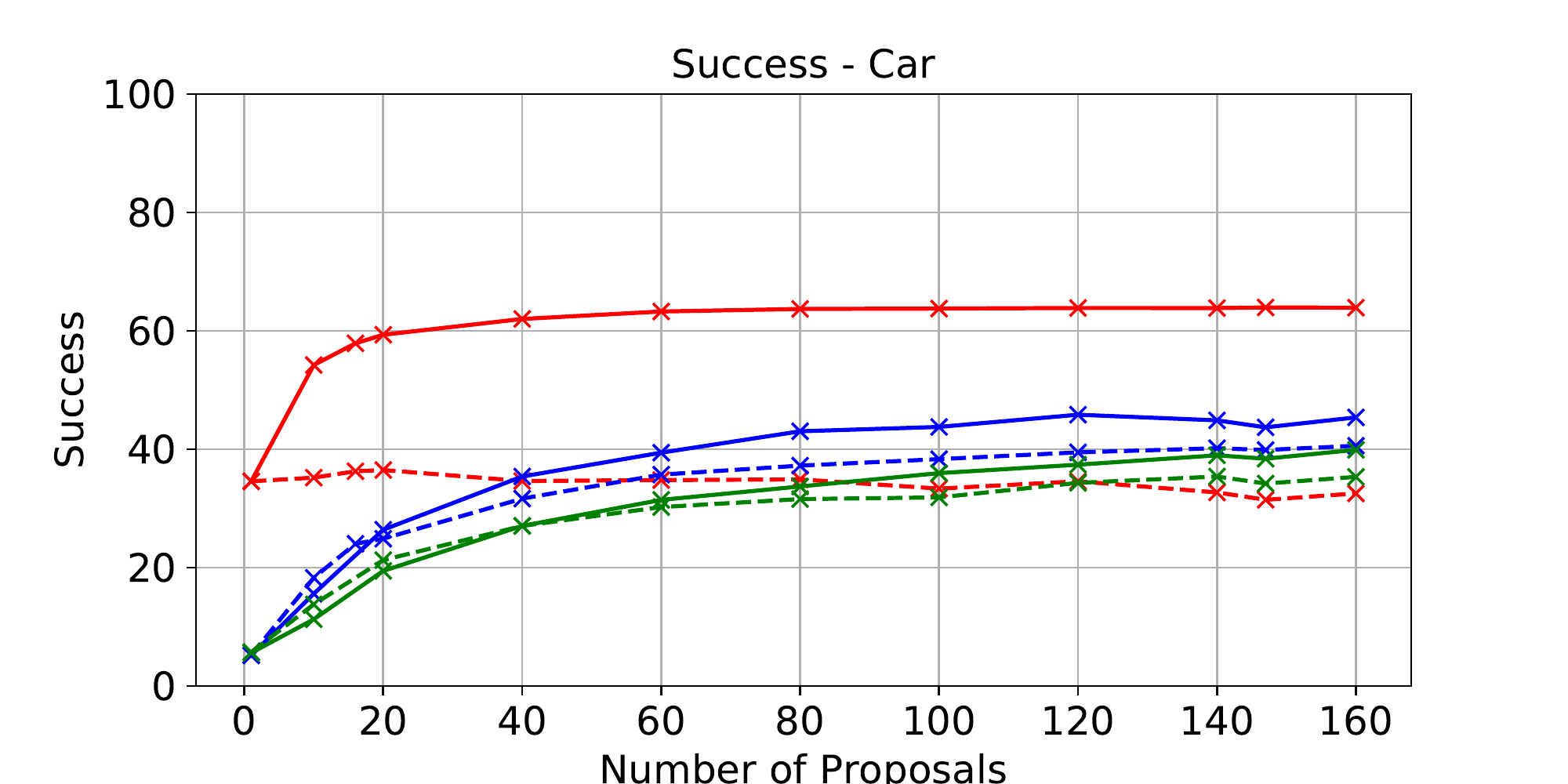}\\
    \includegraphics[width=\textwidth]{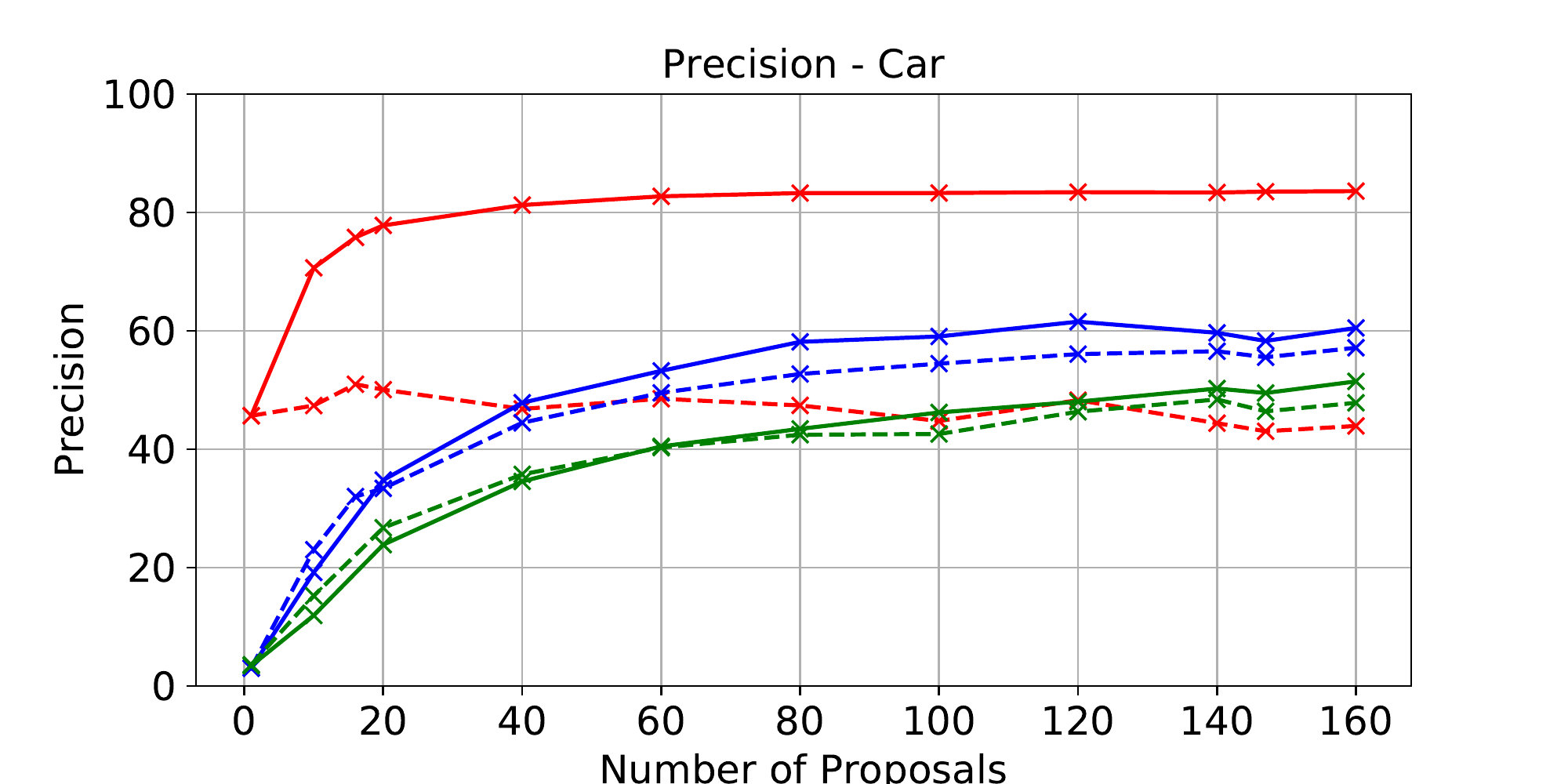}
    \end{minipage}
    \begin{minipage}{0.63\linewidth}
    \centering
	\begin{tabular}{l||c|c}
        Search          & 3D Siamese    & Best proposal \\ \midrule
        KF (top-147)    & \bfseries 41.3 / 57.9   & 43.7 / 58.3 \\ 
        PF (top-147)    &           34.2 / 46.4   & 38.4 / 49.5 \\ 
        % GMM(k= 25)      &           35.6 / 49.1   & 37.9 / 49.3  \\ 
        \midrule
        RPN (top-1)     &           34.6 / 45.7   &  34.6 / 45.7 \\ 
        RPN (top-16)    & \itshape \bfseries36.3 / 51.0   & \itshape \bfseries 57.9 / 75.8 \\ 
        RPN (top-147)   &           29.1 / 40.9   & \bfseries 64.0 / 83.5 \\ 
	\end{tabular}
    \end{minipage}
    \caption{
    \textbf{Left:}
    Proposal results for the class Car using RPN (red), Kalman Filtering (blue) and Particle Filtering (green).
    Continuous lines depict the upper bound results by picking the best proposal ($\simeq$ Recall), while dashed lines depict the discrimination with the 3D Siamese network.
    \textbf{Right:}
    OPE Success and Precision for different Search Space. 
    \textbf{Best} and \textbf{\textit{second}}.
    }
    \label{fig:RPNvsOthers}
\end{figure}

\Figure{RPNvsOthers} depicts a comparison for our Region Proposal Network module (red) to generate proposals over the Kalman Filter (blue) and Particle Filters (green) proposed by Giancola~\etal~\cite{CVPR19}.
We plot the Success and the Precision over the number of proposals considered in both RPN (ours), Kalman Filtering and Particle Filtering. % and Gaussian Mixture Models.
The table report results for specific number of proposals.

\mysection{Proposals Quality}
The continuous lines show the performances when the best candidates are picked among the list of proposals. 
Such results depict an ideal case where the later 3D Siamese Network would be able to pick the bast candidate.
In other words, these are the expected upper bounds when a perfect selection method over the candidates is assumed.
As shown, the best candidate shape from the Region Proposal Network provides significantly better Success and Precision than the best candidate for the Kalman Filtering and Particle Filtering solution.
It appears that RPN proposals saturate after $40$ candidates while the Kalman and Particle Filters are not saturated yet even after $160$ candidates.
While Giancola~\etal~\cite{CVPR19} leverage $147$ candidates for their baseline, we only need around $10$ candidates to outperforms Particle Filtering, which could provide a $~15\times$ speed up for inference in the 3D Siamese similarity selection.
We argue that the Region Proposal Network makes use of the visual features available from the BEV to propose meaningful candidates.
In contrast, the other proposed methods are only relying on a probability distribution and hence are only able to learn motion priors.

\mysection{RPN 3D Siamsese Tracking}
The 3D Siamese network is used on top of the Region Proposal Network in order to discriminate the best candidate among the proposals.
We plot with dashed lines the Success and Precision when leveraging the 3D Siamese tracker for similarity selection.
The number of candidates selected from the BEV Siamese proposal module influences the final tracking results.
Using too few candidates gives more confidence to the BEV Siamese proposals, with $topk=1$ being the equivalent of relying exclusively on the BEV Siamese proposal network.
On the contrary, using too many proposals in inference could confuse the latter 3D Siamese module, not able to pick the best one. As an example, using $147$ proposals provides worse tracking performances than using $16$.
% Generally, it appears that the 3D Siamese network is not able to select the best candidate from the list of proposals.
Generally, it appears that the BEV is able to provide more valuable proposals than KF and PF, but the 3D Siamese network has more difficulty in picking the best among them.

\subsection{Generalization to more Objects}

\begin{figure}[h]
    \centering
    
    \resizebox{\textwidth}{!}{%
    \begin{tabular}{cc|c|c||c|c|c||c|c|c||c|c|c}  &
    \multicolumn{6}{c||}{\textbf{3D Siam - Cyclist}} &
    \multicolumn{6}{c}{\textbf{3D Siam - Pedestrian}} \\ \cmidrule{2-13} &
    \multicolumn{3}{c||}{Success} &
    \multicolumn{3}{c||}{Precision} &
    \multicolumn{3}{c||}{Success} &
    \multicolumn{3}{c}{Precision} \\ \cmidrule{2-13} &
    \textbf{Exh.} & \textbf{\color{blue}KF-147} & \textbf{\color{red}RPN-20} & 
    \textbf{Exh.} & \textbf{\color{blue}KF-147} & \textbf{\color{red}RPN-20} & 
    \textbf{Exh.} & \textbf{\color{blue}KF-147} & \textbf{\color{red}RPN-20} & 
    \textbf{Exh.} & \textbf{\color{blue}KF-147} & \textbf{\color{red}RPN-20} \\ \cmidrule{2-13} &
    \textit{86.91} & {41.53} & \textbf{43.23} &  
    \textit{99.81} & {70.44} & \textbf{81.15} & 
    \textit{71.38} & \textbf{18.24} & {17.89} & 
    \textit{80.43} & {37.78} & \textbf{47.81} \\ \cmidrule{2-13}
    \end{tabular}
    }
    
    \includegraphics[width=0.24\textwidth,trim={1.4cm 0cm 2cm 0.6cm},clip]{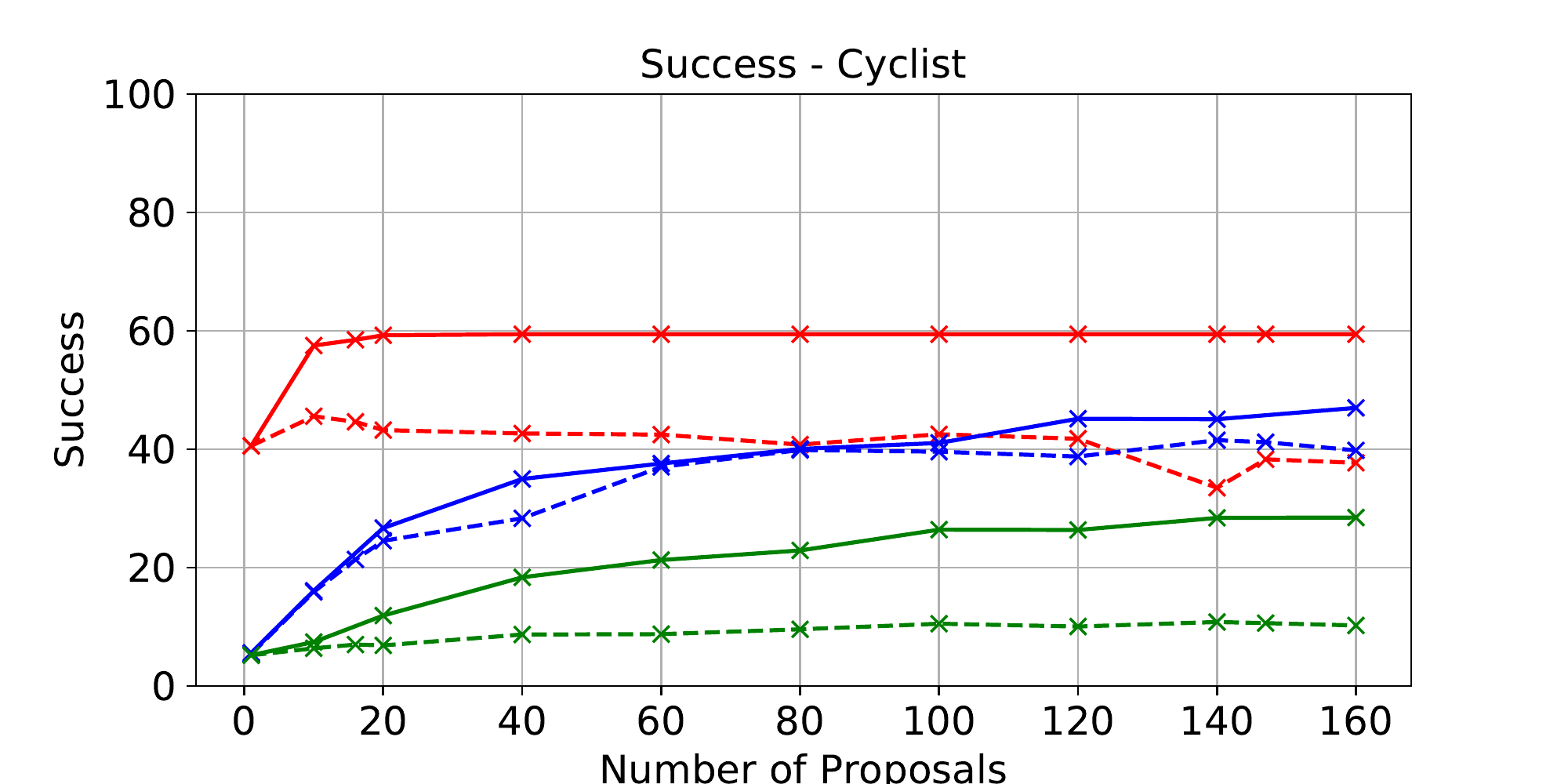}
    \includegraphics[width=0.24\textwidth,trim={1.4cm 0cm 2cm 0.6cm},clip]{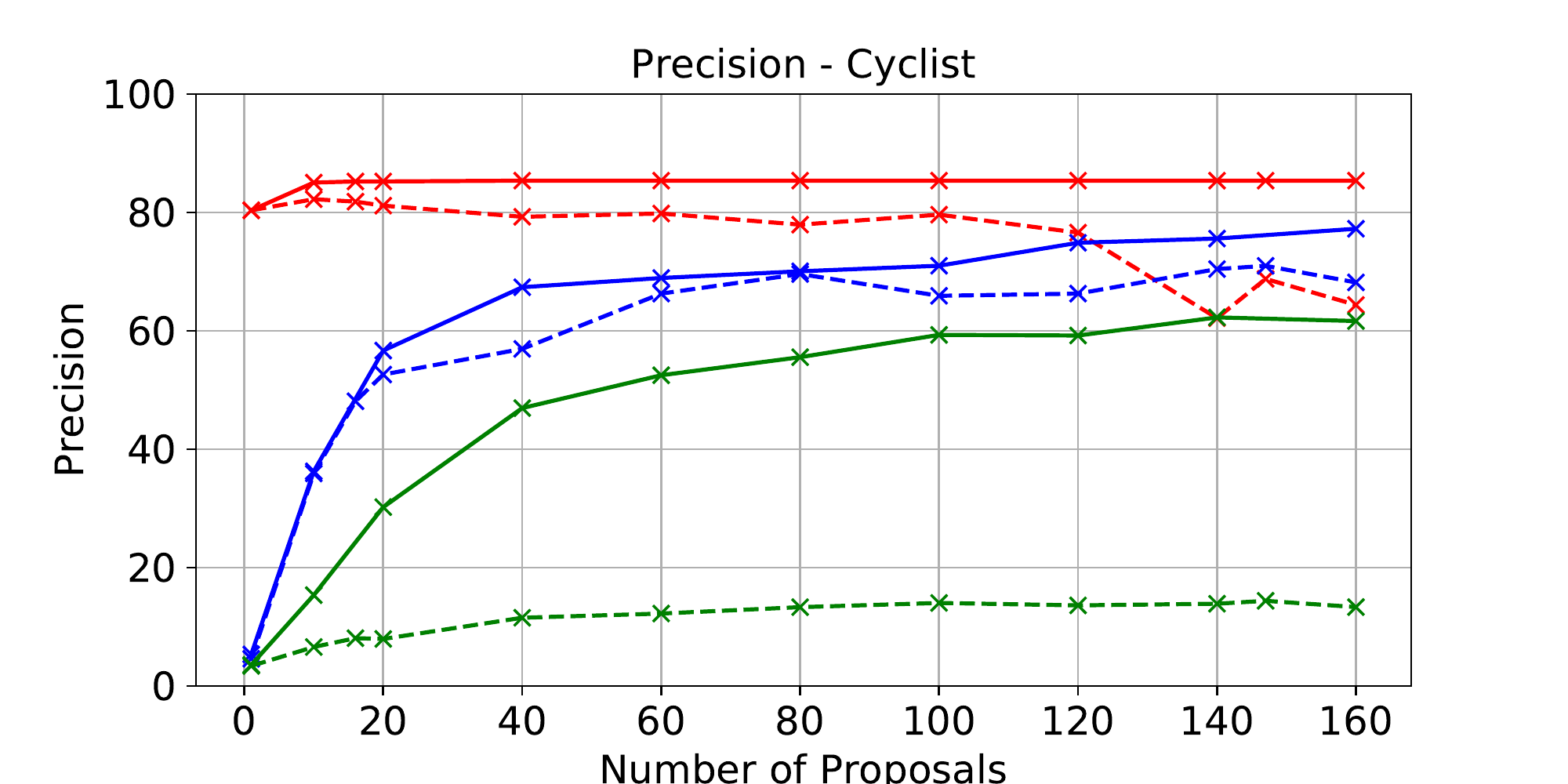}
    \includegraphics[width=0.24\textwidth,trim={1.4cm 0cm 2cm 0.6cm},clip]{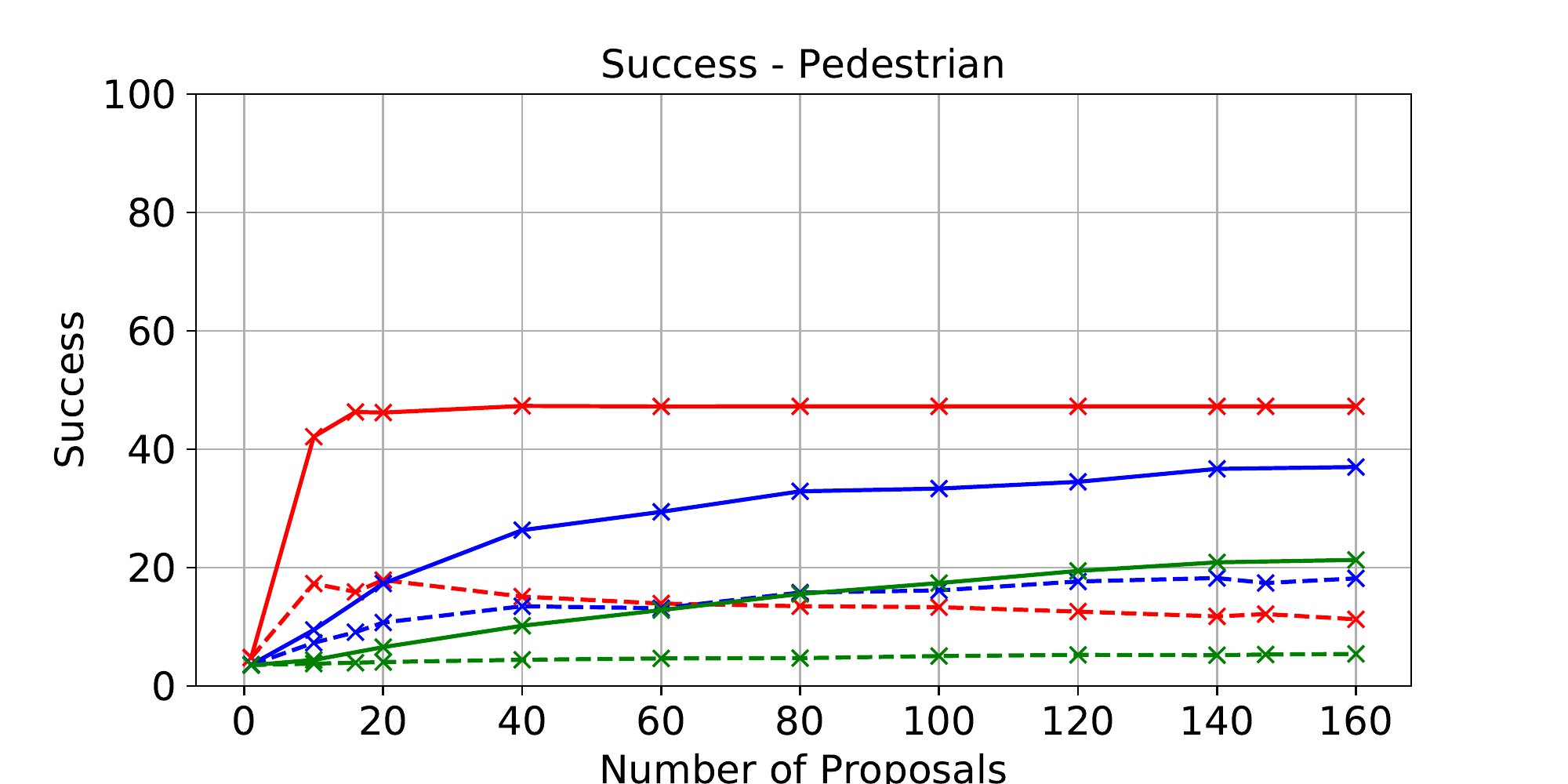}
    \includegraphics[width=0.24\textwidth,trim={1.4cm 0cm 2cm 0.6cm},clip]{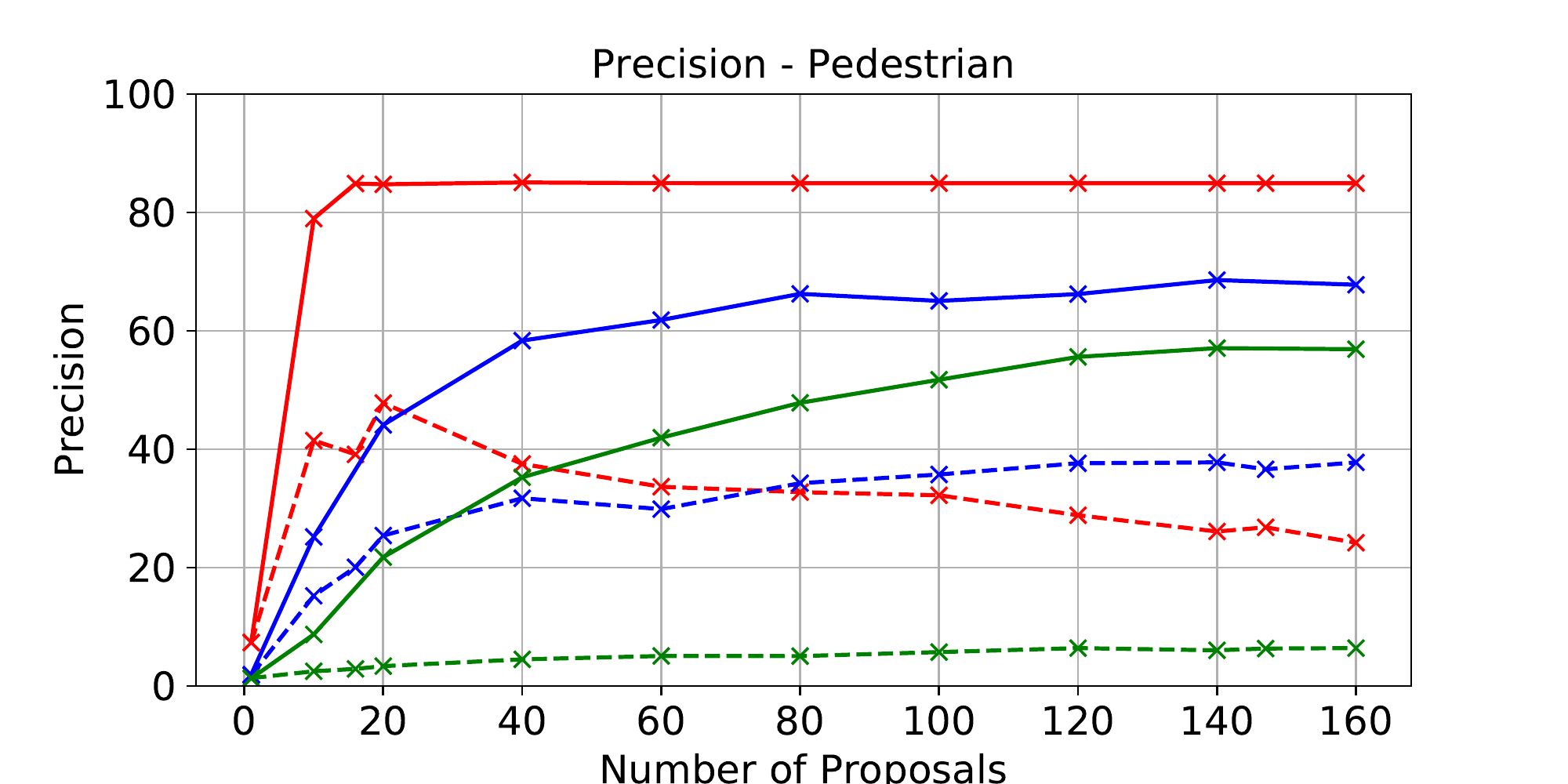}
    
    % \end{adjustbox}
    % \begin{tabular}{l||c|c||c|c} 
                % & \multicolumn{2}{c||}{Cyclist}  &  \multicolumn{2}{c}{Pedestrian} \\ %\cline{2-5} 
                % & Success      & Precision       & Success      & Precision \\
    % \midrule
    % Completion only            (Exhaustive) & 74.00 & 86.72 & 55.61 & 63.25 \\ \hline  
    % Tracking only              (Exhaustive) & 85.66 & 97.99 & \bfseries 74.77 & \bfseries 83.77 \\ \hline   
    % $\lambda_{comp}$@$1e^{-6}$ (Exhaustive) &  86.91 &  99.81 & 71.38 & 80.43 \\ \midrule
        % & 41.53 & 70.44 &  18.24 & 37.78 \\ \hline
        % &  43.23 &  81.15 & 17.89 &  47.81 \\ \midrule
    % \end{tabular}\hfill
    % \end{adjustbox}

    \caption{Proposals for Cyclist and Pedestrian classes using RPN proposals (red), Kalman Filtering (blue) and Particle Filtering (green).
    Continuous lines depict the upper bound results by picking the best proposal while the dashed lines depict our 3D Siamese network.
    }
    \label{fig:RPNvsOthersCyclistPedestrian}
\end{figure}

In this section, we show the generalization capability of our network architecture to different classes of objects such as Cyclists and Pedestrians.
Both classes are deformable by nature, however, in the KITTI dataset, they are considered to be enclosed within bounding boxes which do not vary in size with time.
They can thus be analyzed with our framework without any modification.
We display in \Figure{RPNvsOthersCyclistPedestrian} the performances for our Region Proposal Network against the idealistic Exhaustive Search and more realistic Kalman Filtering proposed by Giancola~\etal~\cite{CVPR19}. The plots are similar to \Figure{RPNvsOthers}, while the above table presents quantitative results for specific top-k proposals.

%
%

%\mysection{Cyclist} In KITTI, the object cyclist is a rigid body that 

\mysection{Region Proposal Network}
We plot with a continuous line the performances of the sole proposal network, by picking the best candidate among $C$ proposals from 
RPN (red), Kalman Filtering (blue) and Particle Filtering (green).
It appears that the proposal network outperforms the Kalman and Particle Filters in both cases, which means that the BEV representation provides enough information to generate meaningful candidates.
Still, the selection from the Siamese BEV is confused between the top-20 proposals, but the 3D Siamese network is able to cope with the fine selection.

% The continuous line in the plot illustrates the best performances how our 

% We highlight here the efficiency of the region proposal network for the additional classes.
% \Figure{RPNvsOthersCyclistPedestrian} illustrates the proposal performance of Kalman Filtering, Particle Filtering and our Region Proposal Network, respectively on Cyclist and Pedestrian classes.
% The continuous lines show the results when picking the best candidate among the set of $C$ proposals, while the selection made by the 3D Siamese network is presented with dashed lines.
%
% It appears that the proposal network outperforms the Kalman and Particle Filters in both cases, which means that the BEV representation provides enough information to generate meaningful candidates.
% Still, the selection from the Siamese BEV is confused between the top-20 proposals, but the 3D Siamese network is able to cope with the fine selection.
% Also, the BEV projection of the Pedestrian is smaller than for the Car and the Cyclist, which makes the BEV representation of Pedestrians less informative and more challenging than for Car and Cyclist.

\mysection{3D Siamese Discriminator}
Similarly, we highlight in dashed line the realistic results using the 3D Siamese tracker to pick the best candidate.
Our BEV-3D Siamese network provides better tracking performances by leveraging less candidates than Kalman and Particle Filtering.
The table above present quantitative results for RPN (ours) and KF-147 \cite{CVPR19}.
Not only do we improve tracking performances, but we improve them by leveraging less candidates, which implies time and complexity improvements as well for the 3D Siamese Tracker.
\subsection{Further insights}
\label{sec:Discussion}
In this section, we provide further insights on our methodology for vehicle tracking. 

% comment on the results

\mysection{2D vs 3D tracking}
The 2D BEV provides greedy information to recognize the position of the vehicle in time.
However, the BEV Proposal module is not able to identify a correct ranking over the other top-k candidates.
% However, the BEV Proposal module is not able to correctly c
As a results, the BEV Proposal module is not able to pick the optimal candidate.
We believe this is due to the loss of information occurring in the BEV representation.
As a results, leveraging the 3D Point cloud similarity selection with the BEV Siamese network improves drastically the tracking performances for all Car, Cyclist and Pedestrian.
% Also, note that the BEV projection of the Pedestrian is smaller than for the Car and the Cyclist, which makes the BEV representation of Pedestrians less informative and more challenging than for Car and Cyclist.

\mysection{Angle regression}
We tried regressing the orientation of the bounding box as well but did not note any improvement.
We argue that the anchors already provided a good enough resolution of $2.5$ degrees without regression, and regressing a value under that resolution is not significant.
Adding an angle regression only increases the method's complexity and induces more errors during inference.

% the higher candidates from the BEV Siamese network is not the bast avaible to pick the bestthe best candidate does not provide enough 
% We can see a
% Yes. Maximum topk 1 is less than 3D with topk16

\mysection{Deformable object}
We noticed with the Pedestrian class that deformable objects do not improve with the shape completion loss (\textit{\textbf{suppl. material}}).
We argue that our model does not take into consideration the shape deformations that occur when a person is walking.
However, leveraging a human body pose encoding could help understanding the complex Pedestrian shape.
It would be possible to extend our methodology this way to include objects outside of the rigid body assumption, but leave that for future work.
% As shown in the experiments section it is possible to achieve competitive results without any modifications on the pipeline.
% However, this could be further improved by adding a pose completion module to account for deformations in objects.
% Also, note that the BEV projection of the Pedestrian is smaller than for the Car and the Cyclist, which makes the BEV representation of Pedestrians less informative and more challenging than for Car and Cyclist.

\mysection{Fast convergence}
We noticed a fast convergence of our end-to-end training, with only a few epochs.
Both networks were pre-trained on ImageNet and KITTI, fulfill two very distinct tasks that only slightly complete each other and thus not much joint training is required.

% We argue that both BEV proposal method and 3D point cloud tracking were pre-trained respectively on ImageNet and KITTI, and thus not much training is required to obtain a good model. 
% Both networks fulfill two very distinct tasks that only slightly complete each other and thus joint training is not required.

\mysection{Gap between proposals and tracking for RPN}
The Region Proposal Network provides a huge improvement in proposals generation with respect to Kalman and Particle Filtering. However, the 3D Siamese network is not able to cope with those candidates since there is a large gap in performance between the upper bounded results and those obtained with the 3D Siamese network.
We believe this gap could be reduced by leveraging a variable variance for the Gaussian distance regression in the 3D Siamese network.

%\mysection{3D regression}
%We tried regressing bounding boxes using 3D information instead of 2D. 
%However regressing in 3D in more complex than in 2D, in particular because of the absence of information in the point cloud.
% general comment on our results

% \newpage
\section{Conclusion}

In this work, we emphasized the importance of searching for tracking.
We leveraged the bijection between 2D BEV and 3D LIDAR bounding in vehicle tracking settings to efficiently generate proposals by searching in 2D.
We used an RPN to generate proposals from LIDAR BEVs which are then discriminated using 3D point cloud information.
We showed that RPNs are more efficient that Kalman and Particle Filters by a large margin.
We then showed that using those proposals helps in selecting the optimal candidates with the 3D Siamese network for vehicle tracking.
We did not only test that for Cars but also for the classes Cyclist and Pedestrian showing consistent improvements in performances.
This method could be extended in the future to perform fine tuning on the selected candidate box through the use of more detailed information contained in 3D point clouds.
However, it is clear that a rough search on 2D BEVs could provide outstanding improvements on tracking performances.

\bibliography{egbib}

\newpage

\section{Supplementary: Exhaustive Search on more Objects}

In this section, we show the tracking performances for Pedestrian and Cyclist with an exhaustive search as per ~\cite{CVPR19}. 
\Table{AblationCyclist} shows that training for shape completion in the ideal case of an exhaustive search improves the tracking performances for the Cyclist and Pedestrian classes. 
We believe that Cyclist can be considered as rigid in time, in particular, the motion of a the biker is negligible for the comprehensive shape.
On  the contrary, Pedestrians may deform significantly which impedes the construction of a reliable model shape representation.
As a result, the completion loss has a negative effect for the final tracking performances.

\begin{table}[tbh]
	\centering
	\caption{Ablation study for the Cyclist class using an exhaustive search space.
    Best results shown in bold.}
	\label{tab:AblationCyclist}
	\begin{tabular}{l||c|c||c|c} 
                & \multicolumn{2}{c||}{Cyclist}  &  \multicolumn{2}{c}{Pedestrian} \\ \cline{2-5} 
                & Success      & Precision       & Success      & Precision \\
    \midrule
    Exhaustive -- Completion only             & 74.00 & 86.72 & 55.61 & 63.25 \\ \hline  
    Exhaustive -- Tracking only               & 85.66 & 97.99 & \bfseries 74.77 & \bfseries 83.77 \\ \hline   
    Exhaustive -- $\lambda_{comp}$@$1e^{-6}$  & \bfseries 86.91 & \bfseries 99.81 & 71.38 & 80.43 \\ \midrule
    \end{tabular}
\end{table}

\section{Supplementary: Point cloud aggregation}

In this section, we are testing our inference architecture (\Figure{Inference}) using different model update methods. 
Since point clouds are easy to combine by concatenating the points, we study what is the best way to create the model point cloud and BEV model in our pipeline.
The model is created dynamically before each frame as a function of the previous results so far. 
In that setting, we are not using the future knowledge nor the ground truth but only the tracking results obtained so far. 
We test the effects of representing the model with only the previous result, only the first frame, a concatenation of the first and previous results, and as a concatenation of all previous tracking results.

As shown in \Table{Aggregation}, it appears that the more frames are used to generate the object model, the better the performances for tracking that object.
Instinctively, we argue that storing more information about the 3D shape of an object obtained from different points of view enhances its capacity to be tracked.
Note that the previous frame may drift apart and not follow the vehicle anymore after some bad results.
Also, the first frame may only contain a few points that are not discriminative enough to describe a vehicle.
However, leveraging a shape completion regularization enhances the semantic information of the 3D latent vector leveraged in the cosine similarity.

\begin{table}[htb]
	\centering
	\caption{OPE Success/Precision for different aggregation fro the Point cloud and BEV models.
    Best representation aggregation shown in bold.}
	\label{tab:Aggregation}
	\begin{tabular}{l||c|c|c} 
Model           & top-147 & top-16 & top-1 \\
% Patch               &    &     & \\
\midrule
Prev. only          & 29.5 / 38.6               & 35.0 / 47.6           & 27.3 / 37.9 \\ \hline
$1^{st}$ only       & 29.9 / 38.3               & 35.7 / 48.4           & 27.3 / 37.9 \\ \hline
$1^{st}$ \& prev.   & 33.2 / 44.3               & 36.4 / 49.6           & 27.3 / 37.9 \\ \hline
All                 & \bfseries 33.9 / 44.8     & \bfseries 37.0 / 49.7 & \bfseries 27.3 / 37.9\\ \hline
    \end{tabular}
\end{table}

\end{document}